# AFSD-Physics: Exploring the governing equations of temperature evolution during additive friction stir deposition by a human-AI teaming approach

Tony Shi[a,b,1], Mason Ma[a,b], Jiajie Wu[a,b], Chase Post[a,b], Elijah Charles[b], Tony Schmitz[b,c]

[a]Systems and Artificial Intelligence Laboratory, University of Tennessee, Knoxville, TN 37996, USA
[b]Machine Tool Research Center, University of Tennessee, Knoxville, TN 37996, USA
[c]Manufacturing Science Division, Oak Ridge National Laboratory, Oak Ridge, TN 37830, USA

**Abstract**

This paper presents a modeling effort to explore the underlying physics of temperature evolution during additive friction stir deposition (AFSD) by a human-AI teaming approach. AFSD is an emerging solid-state additive manufacturing technology that deposits materials without melting. However, both process modeling and modeling of the AFSD tool are at an early stage. In this paper, a human-AI teaming approach is proposed to combine models based on first principles with AI. The resulting human-informed machine learning method, denoted as AFSD-Physics, can effectively learn the governing equations of temperature evolution at the tool and the build from in-process measurements. Experiments are designed and conducted to collect in-process measurements for the deposition of aluminum 7075 with a total of 30 layers. The acquired governing equations are physically interpretable models with low computational cost and high accuracy. Model predictions show good agreement with the measurements. Experimental validation with new process parameters demonstrates the model's generalizability and potential for use in tool temperature control and process optimization.

## 1. Introduction

The goal of learning for either human or artificial intelligence (AI) is to acquire knowledge. Human learning provides a blueprint for machine learning (ML) that is commonly regarded as a pathway to AI. The primary distinction between human and AI learning is that, while humans tend to make inferences

---

about natural laws after acquiring existing knowledge, current AI/ML capabilities perform the inverse: they fit the data generated by embedded laws with little to no awareness of first principles. This lack of awareness limits the use of AI/ML for many science and engineering fields where first principles and physical constraints must be applied. These governing laws have been studied for centuries and are well-known by humans. Therefore, humans tend to focus on the underlying physics. At present, these cognitive capabilities are not integrated within the computational capabilities of AI/ML. With that in mind, this paper aims to advance the knowledge of temperature evolution for additive friction stir deposition (AFSD) by a human-AI teaming approach.

AFSD is an emerging solid state metal additive manufacturing (AM) technology which provides an alternative to fusion-based AM technologies where materials are melted locally using a high-intensity heat source [1, 2]. In the AFSD process, no melting occurs and the geometry and microstructure are produced during layer-by-layer severe plastic deformation. The deposited materials have lower porosity, reduced thermal gradients, lower residual stress, and homogenous microstructure. Prior research efforts have studied various materials and the corresponding structure-property relationships [3-14]. AFSD, together with metrology and computer numerically control (CNC) machining, has been integrated as hybrid manufacturing for industrial applications [15-17].

The parameter settings for AFSD determine the friction and plastic deformation at the tool deposition surface, both of which are primary causes of heat generation in the tool and the build. AFSD modeling efforts are at an early stage. For process modeling, numerical models including both mesh-free methods [18-20] and mesh-based methods [3, 21-24] have been studied. However, multi-physics models are faced with limitations including boundary instability, high computational cost, and poor scalability for multi-layer depositions. These numerical models are often treated as deterministic and are not well-suited to combination with data for learning and accuracy improvement. Further, in-process measurements may be performed to record physical variables like temperature, spindle speed, spindle torque, tool location, and actuator force applied to the feedstock. These in-process measurements reflect unknown physical laws and provide a foundation for the development of ML models. A recent study presented a physics-informed ML method based on neural networks for temperature modeling of the tool and deposition [25]. The neural networks model used in-process measurements with low computational cost but lacked physical interpretability and model generalizability.

Currently, experimentally verified governing equations are not available that can describe the mathematical relationship between temperature evolution at the tool-deposition and other measurable in-process physical variables. The research objective of this paper is to explore the governing equations of temperature evolution during AFSD by a human-AI teaming approach. A human-informed machine learning method, or AFSD-Physics, is proposed to learn the governing equations of temperature evolution at the tool and the build from in-process measurements. Computational simulations and experimental validation are conducted to verify the efficiency and efficacy of the developed methods and acquired governing equations. The primary contributions are:

- The proposed human-AI teaming approach presents a pathway to provide AI with first principles models to advance knowledge in manufacturing. The resulting human-informed machine learning method can effectively explore the unknown physics of AFSD and learn the governing equations of temperature evolution from in-process measurements.

- The acquired governing equations provide physically interpretable robust models with low computational cost and high accuracy.

The paper is organized as follows. In Section 2, the problem description and an overview of the proposed human-AI teaming approach are introduced. Section 3 gives the proposed AFSD-Physics method. The acquired governing equations are discussed in Section 4. The experimental setup and numerical results are presented in Sections 5 and 6. Section 7 provides experimental validation. Section 8 concludes the paper.

## 2. Human-AI teaming for AFSD temperature modeling

### 2.1 Problem description of AFSD temperature evolution

We First principles, human-desired models for AFSD temperature evolution should consider the following physical factors: (1) the thermodynamics phenomena dependent on operating parameters (control variables) for the tool can be formulated as nonlinear dynamic systems with control; (2) temperature evolution during layer-by-layer deposition exhibits thermal cycling as the heating and cooling steps alternate; and (3) governing equations with high-accuracy and low computational cost enable real-time prediction, process control, and optimization.

The Eq. (1) ordinary differential equation (ODE) provides the governing equation for temperature evolution in this paper:

$$\dot{T} = f(T, \boldsymbol{u}), \qquad (1)$$

where $T$ can be either the tool temperature $T_{tool}$ or the build temperature $T_{build}$, $f(\cdot)$ is an unknown function that governs the dynamics of $T$. $\boldsymbol{u} = [\boldsymbol{u}_1\ \boldsymbol{u}_2]$ can include other process variables besides $T$, where $\boldsymbol{u}_1$ refers to the vector of process parameters that can be explicitly selected and controlled, such as tool spindle speed and feedstock and tool feed velocities, and $\boldsymbol{u}_2$ includes process physical variables that cannot be explicitly selected but can be measured, such as spindle torque and actuator force to push the feedstock through the rotating spindle. A thermal cycle consists of a heating and a cooling stage as defined in Eq. (2):

$$P^l = P^l_{heat} \cup P^l_{cool}, \qquad \forall l \in [L], \tag{2}$$

where $[L] \coloneqq \{1, 2, \ldots, L\}$ denotes the set of deposited layers. Since the values of $\boldsymbol{u}$ are different in $P^l_{heat}$ and $P^l_{cool}$, it is plausible to model different functions $f(\cdot)$ for temperature in $P^l_{heat}$ and $P^l_{cool}$, especially for the tool temperature $T_{tool}$.

Next, the in-process physical variables measured for this paper are summarized as follows:

$T_{tool}$:    $T_{tool} = T_{tool}(t)$, tool temperature at time $t$ measured by a tool-embedded thermocouple axially located within 0.25 mm to 0.38 mm of the tool surface. An infrared camera is also attached to the moving spindle carriage. The peak temperature in the field of view is compared with the tool-embedded thermocouple temperature measurement to confirm the performance of the two sensors.

$T_{build}$:    $T_{build} = T_{build}(\boldsymbol{s}, t)$, build/substrate temperature at time $t$ for location $\boldsymbol{s} \in \mathcal{S}$. $\mathcal{S} = \{\boldsymbol{s}_1, \boldsymbol{s}_2, \boldsymbol{s}_3, \boldsymbol{s}_4\}$ refers to a set of four locations where four equally spaced thermocouples were embedded 2.54 mm below the substrate surface along the deposition direction to measure the substrate temperature.

$\omega$:    Tool spindle speed;

$f_t$:    Tool feed velocity or tool traverse speed;

$f_m$:    Feedstock feed velocity or material feed rate;

$T_f$:    Spindle torque applied to the spindle to overcome the friction force;

$P_f$:    Spindle power;

$T_m$:    Servo torque for feedstock material;

$F_m$:    Force applied to the feedstock to push it through the rotating tool;

$\boldsymbol{s}_{tool}$:    $\boldsymbol{s}_{tool} = (s_x, s_y, s_z)$, position of the center point at the tool bottom surface;

$\boldsymbol{v}_{tool}$:    $\boldsymbol{v}_{tool} = (v_x, v_y, v_z)$, velocity of the tool center at the bottom surface along the machine's X, Y and Z axes.

The fundamental research question is: How can the analytical forms of $\dot{T} = f(T, \boldsymbol{u})$ be obtained to describe the temperature evolutions at the tool and the build from in-process measurements? For the tool, the governing equation is desired to describe the heat generation mechanism and temperature evolution based on the AFSD operating parameters. For the build, the governing equation is desired to establish the relationship between temperature measurements at the tool and temperature measurements at the build/substrate.

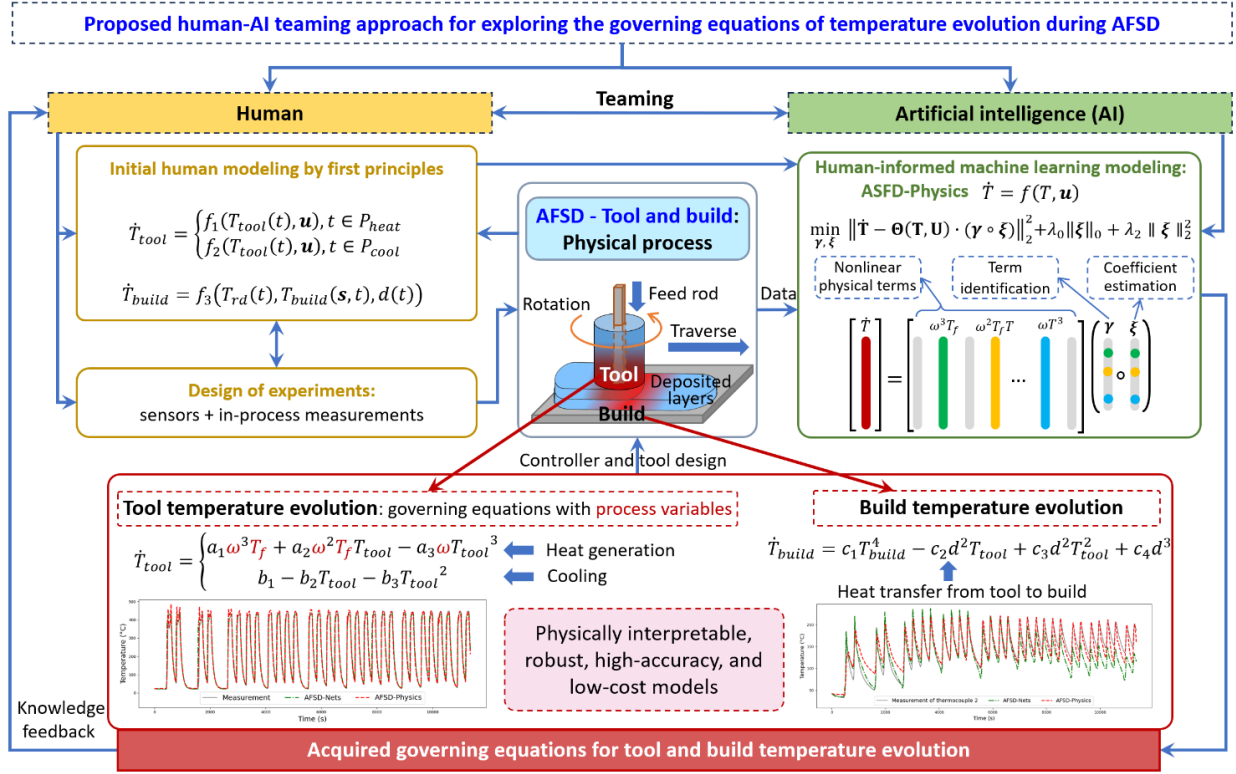

*Figure 1. Schematic illustration of the proposed human-AI teaming approach for AFSD temperature evolution modeling.*

**2.2 Overview of the proposed human-AI teaming approach**

A human-AI teaming approach is proposed to provide the desired AFSD model. The general steps as shown in Figure 1 are summarized.

*Step 1. Human initial human modeling by first principles.*

This first step refers to how the human integrates first principles models and in-process measurements of physical variables with AI (Section 3.1).

*Step 2. Design of experiments for data collection.*

Based on the model requirements, measurability of physical variables and available sensors,

experiments are designed and conducted for a set of process parameters (Section 5).

*Step 3. Human-informed machine learning modeling.*

*3.1 Design of human-informed learning function space*. The learning function space design integrates physical knowledge and augments the human-informed initial model by ML (Section 3.2.1).

*3.2 Design of loss function and optimization algorithm*. With the human-informed learning function space, an effective discrete optimization algorithm is designed to minimize appropriate loss functions with respect to accuracy, time efficiency, and robustness to noise in data (Section 3.2.2).

*3.3 Learning process to acquire governing equations*. Experimental data from in-process measurements is used for the training process of human-informed ML to obtain the governing equations (see Sections 4, 6, and 7).

*Step 4. Experimental validation.*

Analyses of the acquired models, including physical interpretation, simulation, and experimental validation, are conducted. An additional round of Steps 1-3 is performed (see Sections 4, 6, and 7).

In the following sections, the procedure is presented step-by-step to identify the governing equations of temperature evolution during AFSD.

## 3. Proposed AFSD-Physics

This The resulting AFSD-Physics method is presented in terms of its two key components: initial human modeling by first principles and human-informed ML modeling.

### 3.1 Initial human modeling by first principles

Initial human modeling combines the physical process of AFSD with in-process measurements.

*3.1.1 Modeling for tool temperature*

The governing equation for the tool temperature is described by the Eq. (3) piecewise functions for the heating and cooling stages, respectively:

$$\dot{T}_{tool} = \begin{cases} f_1(T_{tool}(t), \boldsymbol{u}), & t \in P_{heat}^l, l \in [L], \quad (3a) \\ f_2(T_{tool}(t), \boldsymbol{u}), & t \in P_{cool}^l, l \in [L]. \quad (3b) \end{cases}$$

Function $f_1(\cdot)$ describes the critical mechanism of heat generation. Vector $\boldsymbol{u}$ in $f_1(\cdot)$ can include process parameters related to both frictional and adiabatic heating, including tool spindle speed, tool traverse speed, feedstock feed velocity, spindle torque, etc. Function $f_2(\cdot)$ describes the tool cooling mechanism when the tool is not in contact with the build. Vector $\boldsymbol{u}$ in $f_2(\cdot)$ may include parameters for controlling the tool cooling jacket which surrounds the rotating tool and provides forced chilled fluid heat rejection, as needed.

The tool temperature model in Eq. (3a) describes the moving heat source. During AFSD, the moving heat source is the deposit beneath the rotating tool, where both friction heating and adiabatic heating occurs. However, the measurements of the deposit beneath the tool are difficult to obtain. The assumption is that the rotating deposit is a single point heat source and its temperature $T_{rd}(t)$ is equal to the tool temperature as shown in Eq. (4),

$$T_{rd}(t) = T_{tool}(t), \qquad (4)$$

where $T_{tool}(t)$ can be measured by an embedded thermocouple, which is radially offset from the tool center (outside the square bore) and is located axially close to the tool surface. As such, the temperature measured by the thermocouple is assumed to be the same as rotating deposit. Note that the actual path of the thermocouple during the rotating-translating deposition tool motion is cycloidal in nature. With this assumption, temperature samples from the cycloidal path of the thermocouple are used as in-process measurements from the linear motion of the single point heat source with constant traverse speed. Equation (4) is used to link heat generation at tool and heat transfer at the build (deposit).

*3.1.2 Modeling for build temperature*

The 3D build temperature evolution is naturally governed by a heat transfer partial differential equation (PDE) across the build geometry. However, in-process measurements are only available for the four thermocouples embedded in the substrate (build plate) at discrete; see Section 5 for the details. This significantly limits the information available for AI modeling. In this first attempt, the ordinary differential equation (ODE) that incorporates the temperature of and distance to the moving heat source is considered to model the build temperature as shown in Eq. (5),

$$\dot{T}_{build} = f_3\big(T_{rd}(t), T_{build}(\boldsymbol{s}, t),\, d(t)\big), \qquad t \in P_{heat}^l \cup P_{cool}^l,\ l \in [L], \qquad (5)$$

where $T_{build} = T_{build}(\boldsymbol{s}, t)$ is the build temperature at location $\boldsymbol{s} \in \boldsymbol{S}$, and $\boldsymbol{S} = \{\boldsymbol{s}_1, \boldsymbol{s}_2, \boldsymbol{s}_3, \boldsymbol{s}_4\}$ refers to a set of four locations where the four equally spaced thermocouples were embedded in the substrate along the deposition direction to measure the substrate temperature. $d(t)$ is the Euclidean distance between the point location of the rotating deposit $\boldsymbol{s}_{rd}$ and arbitrary location $\boldsymbol{s} \in \boldsymbol{S}$, i.e., $d(t) = \| \boldsymbol{s}_{rd} - \boldsymbol{s} \|_2$. From the assumption in the previous section, it can be obtained that $\boldsymbol{s}_{rd} = \boldsymbol{s}_{tool}$. By introducing $d(t)$, Eq. (5)

implicitly incorporates the spatial information of the build geometry. Therefore, it can be used to describe the spatially distributed temperature for all locations $s \in S$. Note here a single function $f_3(\cdot)$ is adopted for both $P_{heat}^l$ and $P_{cool}^l$.

Finally, the initial conditions for tool temperature and build temperature are provided as shown in Eq. 6:

$$T_{tool}(0) = T_0, T_{build}(s_{tc}, 0) = T_0^{s_{tc}}, \qquad (6)$$

where $s_{tc}$ is the location of the thermocouple in the substrate. In short, Eqs. (3)-(6) collectively provide the initial human modeling for the governing equations of the unknown temperature evolution at the build and tool.

**3.2 Human-informed machine learning modeling**

This section presents the human-informed ML modeling to learn the analytical form of the unknown functions $f_1(\cdot)$, $f_2(\cdot)$ and $f_3(\cdot)$ in initial human modeling by first principles. The main idea is to first design the human-informed learning function spaces for $\dot{T}_{tool}$ and $\dot{T}_{build}$, and then design the loss function and optimization algorithm for the learning process to acquire the governing equations from in-process measurements. The discrete-optimization based machine learning algorithm in [26, 27] is extended for the design of loss function and optimization algorithm.

Let $\mathbf{T}_{tool} \in \mathbb{R}^{N \times 1}$, $\mathbf{T}_{build} \in \mathbb{R}^{N \times 1}$ and $\mathbf{U} \in \mathbb{R}^{N \times J}$ denote the in-process measurements of tool temperature, built temperature at specific location and the other $J$ physical variables, respectively. The in-process measurements are recorded at a total of $N$ time points. Let $\dot{\mathbf{T}}_{tool} \in \mathbb{R}^{N \times 1}$ and $\dot{\mathbf{T}}_{build} \in \mathbb{R}^{N \times 1}$ denote the time derivatives of temperatures, which are numerically calculated by finite difference with smooth techniques.

*3.2.1 Design of human-informed learning function space*

Without loss of generality, the notations for $f_1(\cdot)$, $f_2(\cdot)$ and $f_3(\cdot)$ are not distinguished for the ease of presentation of this section. Let $[P] = \{1,2,\cdots,P\}$ for arbitrary positive integer throughout, where $P \in \mathbb{Z}_+$. A candidate set consisting of $P$ nonlinear physical terms is defined in Eq. (7),

$$\boldsymbol{\theta}(T, \boldsymbol{u}) = [1 \ T \ \boldsymbol{u} \ (T \otimes T) \ (\boldsymbol{u} \otimes \boldsymbol{u}) \ (T \otimes \boldsymbol{u}) \ \cdots \ ], \qquad (7)$$

where operator $\otimes$ denotes the element-wise combinations for constructing nonlinear terms of $T$ and/or $\boldsymbol{u}$. For example, polynomial terms of $\boldsymbol{u}$ and $T$ are included as basic nonlinear physical terms. Other nonlinear physical terms like trigonometric functions can also be added into the candidate set if necessary. Let $\theta_p$ denote the $p$-th term of $\boldsymbol{\theta}(T, \boldsymbol{u})$, $\forall p \in [P]$. In this paper, assuming functions $f_1(\cdot)$, $f_2(\cdot)$ and $f_3(\cdot)$ live in the affine space spanned by $\boldsymbol{\theta}(T, \boldsymbol{u})$, the human-informed learning function space is given in Eq. (8),

$$f(T, \boldsymbol{u}) = \boldsymbol{\theta}(T, \boldsymbol{u}) \cdot \boldsymbol{\xi}, \tag{8}$$

where $\boldsymbol{\xi} = [\xi_1 \ \xi_2 \ \cdots \ \xi_P]^T \in \mathbb{R}^{P \times 1}$ refers to the vector of coefficients. Thus, to identify the governing equations is to estimate the coefficients $\boldsymbol{\xi}$ from in-process measurements. Another assumption for the proposed method is that $\boldsymbol{\xi}$ is sparse for governing equations, i.e., only parsimonious terms that are active govern the dynamics of the underlying physics of AFSD.

*3.2.2 Design of loss function and optimization algorithm*

Based on the human-informed function space, the loss function together with the resulting sparse optimization problem is defined in Eq. (9),

$$\min_{\boldsymbol{\xi}} \left\| \dot{\mathbf{T}} - \boldsymbol{\Theta}(\mathbf{T}, \mathbf{U}) \cdot \boldsymbol{\xi} \right\|_2^2 + \lambda_0 \|\boldsymbol{\xi}\|_0 + \lambda_2 \|\boldsymbol{\xi}\|_2^2, \tag{9}$$

where $\boldsymbol{\Theta}(\mathbf{T}, \mathbf{U}) \in \mathbb{R}^{N \times P}$ is the augmented matrix obtained by evaluating $\boldsymbol{\theta}(T, \boldsymbol{u})$ at each time point of $\mathbf{T}$ and $\mathbf{U}$. The first term represents the empirical risk minimization principle between the approximations and the derivatives of temperature measurements $\dot{\mathbf{T}}$. The regularization term $\|\boldsymbol{\xi}\|_0$ denotes the $\ell_0$-norm (pseudo norm) of $\boldsymbol{\xi}$, namely the number of nonzero terms in $\boldsymbol{\xi}$. The inclusion of the $\|\boldsymbol{\xi}\|_0$ with weight $\lambda_0$ enables the selection of a parsimonious set of $\boldsymbol{\theta}(T, \boldsymbol{u})$ for describing the underlying physics. The regularization term $\|\boldsymbol{\xi}\|_2^2$ with weight $\lambda_2$ is added to reduce the effects of noise in the measurements.

To indicate the existence of terms of $\boldsymbol{\theta}(T, \boldsymbol{u})$ in $f(T, \boldsymbol{u})$, a vector of discrete variables, $\boldsymbol{\gamma} = [\gamma_1 \ \gamma_2 \ \cdots \ \gamma_P]^T \in \mathbb{B}^{P \times 1}$ is introduced, where $\mathbb{B} = \{0,1\}$ is the Boolean domain and

$$\gamma_p = \begin{cases} 1, & \text{if } f(\boldsymbol{u}) \text{ includes } \theta_p \\ 0, & \text{otherwise} \end{cases}, \quad \forall p \in [P]. \tag{10}$$

With $\boldsymbol{\gamma}$, Eq. (9) can be rewritten to indicate the inclusion of each term in $\boldsymbol{\theta}(T, \boldsymbol{u})$ as

$$\min_{\boldsymbol{\xi}} \left\| \dot{\mathbf{T}} - \boldsymbol{\Theta}(\mathbf{T}, \mathbf{U}) \cdot (\boldsymbol{\gamma} \circ \boldsymbol{\xi}) \right\|_2^2 + \lambda_0 \|\boldsymbol{\xi}\|_0 + \lambda_2 \|\boldsymbol{\xi}\|_2^2, \tag{11}$$

where $\boldsymbol{\gamma} \circ \boldsymbol{\xi}$ refers to the element-wise (Hadamard) product. Consequently, a mixed-integer optimization problem is given by reformulating Eq. (11) as

$$\langle \boldsymbol{\gamma}^*, \boldsymbol{\theta}^* \rangle = \underset{\langle \xi, \gamma \rangle \in \Delta}{\operatorname{argmin}} \left\| \dot{\mathbf{T}} - \boldsymbol{\Theta}(\mathbf{T}, \mathbf{U}) \cdot \boldsymbol{\xi} \right\|_2^2 + \lambda_2 \|\boldsymbol{\xi}\|_2^2, \tag{12a}$$

where the solution space $\Delta$ is defined as

$$\Delta \coloneqq \{ \langle \boldsymbol{\xi}, \boldsymbol{\gamma} \rangle \mid -M\boldsymbol{\gamma} \leq \boldsymbol{\xi} \leq M\boldsymbol{\gamma}, \boldsymbol{\gamma}^T \boldsymbol{e} = k, \boldsymbol{\xi} \in \mathbb{R}^{P \times 1}, \boldsymbol{\gamma} \in \mathbb{B}^{P \times 1} \}, \tag{12b}$$

where $M$ is a constant number and can be identified from data. $[-M, M]$ defines the lower and upper bounds of $\boldsymbol{\xi}$. When $\gamma_p = 1$, $\xi_p$ is estimated based on optimization within $[-M, M]$; otherwise, when $\gamma_p = 0$, $\xi_p = 0$, meaning term $\xi_p$ is not included in $f(T, \boldsymbol{u})$. Vector $\boldsymbol{e} \in \mathbb{R}^{P \times 1}$ is a vector with all entries set to 1. As a result, $\boldsymbol{\gamma}^T \boldsymbol{e}$ refers to the number of nonzeros in $\boldsymbol{\gamma}$, which is defined as constant $k$. Parameters $\lambda_2$ and $k$ can be tuned by validation and cross validation techniques (see [26] for details).

A two-stage solution procedure is utilized to solve the problem. First, $\boldsymbol{\gamma}^*$ is identified by solving Eq. (12) with a discrete optimization solver using column-wise normalized data of $\boldsymbol{\Theta}(\mathbf{T}, \mathbf{U})$ and $\dot{\mathbf{T}}$. The normalization removes the effects of different scales of terms in $\boldsymbol{\theta}(T, \boldsymbol{u})$. In the second stage, $\boldsymbol{\xi}^*$ is estimated by a least squares algorithm using the original data $\boldsymbol{\Theta}(\mathbf{T}, \mathbf{U})$ and $\dot{\mathbf{T}}$. Only columns corresponding to $\boldsymbol{\gamma}^*$ are used for the estimation. This allows coefficients $\boldsymbol{\xi}^*$ to represent the inherent quantitative relationship between the selected terms in $\boldsymbol{\theta}(T, \boldsymbol{u})$ and $\dot{T}$.

Once $\boldsymbol{\gamma}^*$ and $\boldsymbol{\xi}^*$ are identified, the acquired governing equation can be expressed as

$$\dot{T} = \boldsymbol{\theta}(T, \boldsymbol{u}) \cdot (\boldsymbol{\gamma}^* \circ \boldsymbol{\xi}^*). \tag{13}$$

The acquired governing equations, as knowledge feedback, can inform humans to conduct further analyses, including physical interpretation, simulation, and experimental validation. Note that multiple-round teaming between humans and AI would be needed in a closed-loop manner.

## 4. Main results: Acquired governing equations

The acquired governing equations by AFSD-Physics method are presented here. For details of the learning procedure, readers are referred to the collection of in-process measurements by experiments in Section 5 and the numerical experiment settings for AFSD-Physics in Section 6. With such settings, AFSD-Physics delivers all the governing equations in Eqs. (14) and (15) within 30 seconds.

The acquired governing equation for governing the evolution of tool temperature $T_{tool}$ (°C) is

$$\dot{T}_{tool} = \begin{cases} a_1\omega^3 T_f + a_2\omega^2 T_f T_{tool} - a_3\omega T_{tool}^3, t \in P_{heat}, & (14a) \\ b_1 - b_2 T_{tool} - b_3 T_{tool}^2, \quad t \in P_{cool}, & (14b) \end{cases}$$

where $\omega$ is the tool spindle speed (rpm), $T_f$ is the spindle torque (Nm) required to overcome the friction force, $\boldsymbol{a} = [a_1\ a_2\ a_3]$ and $\boldsymbol{b} = [b_1\ b_2\ b_3]$ are the estimated coefficients. As suggested by previous studies [12, 18], spindle speed, feedstock feed velocity, and the force experienced by the tool are the major factors for heat generation. Equation (14a) describes the heat generation mechanism with respect to the spindle speed $\omega$ and the spindle torque $T_f$. At present, force measurements are not available due to the lack of effective metrology methods, but the inclusion of $T_f$ indicates the significance of force given the inherent relation between force and torque. Since $T_f \omega$ refers to the friction power, terms $\omega^3 T_f$ and $T_{tool}\omega^2 T_f$ may indicate unknown underlying heat generation mechanisms in addition to friction power. Note that the feedstock feed velocity $f_t$ is included in the human-informed learning function space $\boldsymbol{\theta}(T, \boldsymbol{u})$. However, the human-informed machine learning modeling does not select any terms containing $f_t$. The reason may be because $f_t$ in current experiments is a constant at 1.93 mm/second during deposition.

For the governing equation of tool cooling shown in Eq. (14b), it is observed that the first two terms exactly match Newton's law of cooling, with constant $b_1$ relevant to the room temperature. In addition, a new term $-b_3 T_{tool}^2$ is acquired. For the cooling stage, $\dot{T}_{tool}$ is always non-positive. The negative coefficient $-b_3$ indicates the temperature decrease of the AFSD tool during cooling is much faster than that described by Newton's cooling law. This may represent the additional cooling effect from the external water jacket.

The governing equation for governing the evolution of build temperature $T_{build}$ (°C) is presented in Eq. (15).

$$\dot{T}_{build} = c_1 T_{build}^4 - c_2 d^2 T_{tool} + c_3 d^2 T_{tool}^2 + c_4 d^3, \quad t \in P_{heat} \cup P_{cool}, \quad (15)$$

where distance $d(t) = \|\boldsymbol{s} - \boldsymbol{s}_{rd}(t)\|_2$ (mm) and $\boldsymbol{c} = [c_1\ c_2\ c_3\ c_4]$ is the vector of coefficients. Equation (15) can be described as an ODE with time variant parameter $d(t)$. Also, the spatial information of the location $\boldsymbol{s} \in \boldsymbol{S}$ in the substrate is implicitly included in $d(t)$. As such, Eq. (15) can predict the temperature profile for all points in the centerline of the substrate where the thermocouples are located. This will generate a 1D map for temperature distribution. In addition, based on the assumption of a single point moving heat source, Eq. (15) may have the potential to predict the temperature evolution at the 2D vertical plane of the deposited layers bounded by the centerline of the bottom layer and the linear motion path of the single point heat source.

With Eqs. (14)-(15), time domain simulation can be conducted for a multi-layer AFSD process. In general, two types of time domain simulations based on different initial value settings are considered.

- *Type I simulation*: A new initial value for each layer of simulation. Type I simulation is used for simulating the heat generation in Eq. (14a) and tool cooling in Eq. (14b) independently. The measurement at the beginning of each layer is used to set the initial value. This represents the scenario where in-process temperature measurements are available to calibrate the simulation layer-by-layer.

- *Type II simulation*: A single initial value for the entire multi-layer simulation. The initial value of the next layer is set to be the last value of the previous layer. Type II simulation is used for the coupled simulation of heat generation and tool cooling as well as build temperature evolution in Eq. (15). This simulation is practical when no in-process temperature measurements are available and only the room temperature is used to start the simulation.

**5. Experimental setup**

Experiments were performed using a commercially available MELD Manufacturing L3 machine to deposit solid wrought aluminum 7075 feedstock rod with 9.53 mm × 9.53 mm × 508 mm dimensions on the substrate of the same material. The build to be deposited was a wall with a length of 216 mm and a height of 45.6 mm (30 layers with each layer 1.52 mm thickness). As shown in Figure 2(a), the deposition was performed in a single direction and the tool was returned to the same starting position for each layer. Given the wall length, a single rod of wrought feedstock was able to deposit two layers. Then the next rod was manually inserted into the spindle from the square bore at the bottom of the tool.

The operating parameters were: 135 rpm spindle speed during deposition, 115.6 mm/min feedstock feed velocity, and 127 mm/min tool feed velocity (traverse speed along the deposition direction). At the beginning of each layer, the spindle speed was set at 350 rpm to heat the material and was reduced to 135 rpm as the deposition began. A cooling jacket located around the tool was run at a constant flow rate during deposition to cool the tool and avoid excessive heating of the rod within the tool, which can cause the deposition to fail due to adhesion between the rod and tool internal passage. After the numerical results in Section 6 were obtained, additional experiments using a spindle speed of 115 rpm was also performed for validation. See details in Section 7.

In-process measurements were obtained for both the tool and the build. The tool had an embedded K-type thermocouple (KTC), which was located within 0.25 mm to 0.38 mm of the tool surface and used to measure the tool-deposit interface temperature. The L3 controller was also able to record time series data, including spindle speed, feedstock feed velocity and tool position, at a 1 Hz sampling frequency. For the substrate temperature, two baseplates were used. The upper baseplate was the one where the deposit was made, and four equally spaced K-type thermocouples were embedded 2.54 mm below the build surface and along the track direction from the underside of the plate. Figure 2(b) shows the locations of the four thermocouples. The underside in Figure 2(b) was covered by the lower baseplate to protect the thermocouples.

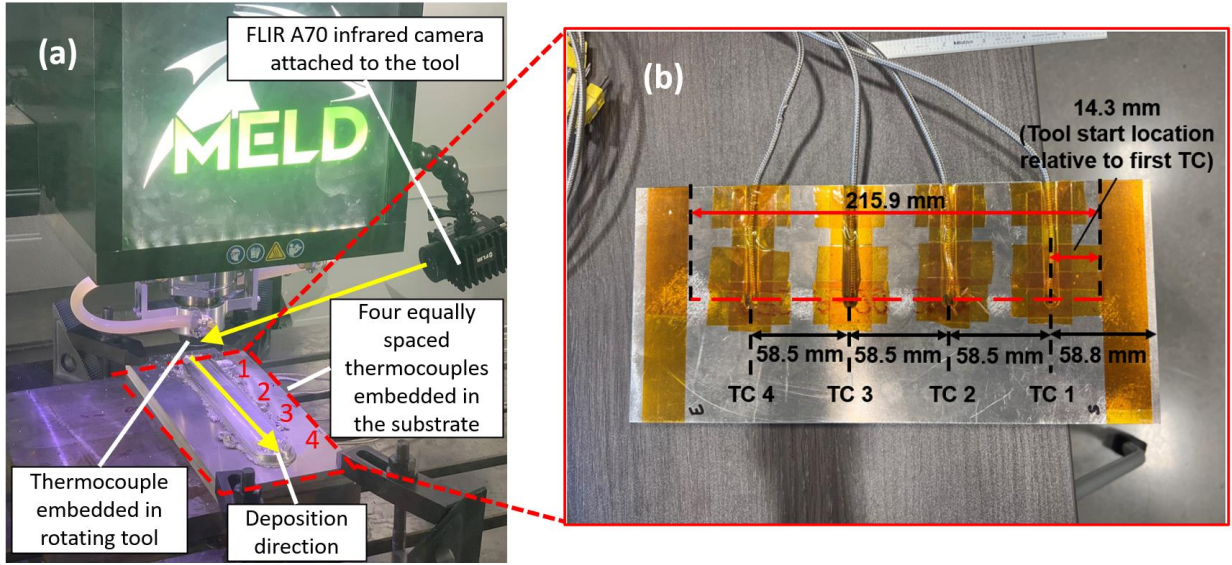

*Figure 2. Experimental setup for depositing 7075 aluminum feed rod on a 7075 aluminum substrate for 30 layers using the MELD Manufacturing L3 machine. (a) A thermocouple embedded in the tool was used to measure the tool temperature. Four equally spaced thermocouples were embedded in the substrate for measuring temperature evolution beneath the deposition track. (b) Substrate with locations of the four thermocouples. Figure is from [25].*

The experiments resulted in 11,358 in-process data points. The features include the L3 controller-captured data (spindle speed, torque, and power; feedstock feed velocity and actuator force and torque; and position, velocity, and torque for the X, Y and Z axes), the tooltip temperature from the embedded thermocouple, and the temperatures from the four substrate thermocouples.

6. **Computational experiments**

The This section presents the numerical results and analysis for the proposed AFSD-Physics method using the in-process measurements, denoted by 135-rpm, from Section 5. The simulation results for the acquired models as described in Section 4 are presented. Comparisons with AFSD-Nets [25] are conducted to show the effectiveness of AFSD-Physics. All computations are completed using Python and executed using a mobile workstation with an Intel® Xeon® W-10885M CPU @ 2.40GHz, 128 GB memory, 64-bit Windows 10 Pro operating system for workstations. The CPLEX 20.1 solver is used as the discrete optimizer in human-informed machine learning modeling to solve the problem in Eq. (12).

### 6.1 Numerical experiment settings of AFSD-Physics

*Training and testing dataset.* Measurements from the 5-th to the 20-th layer of the 135-rpm dataset are used as training data. The first 5 layers are discarded due to poor data quality at the beginning of the AFSD process. The measurements from the last 10 layers of the 135-rpm dataset are used as testing data. Note that for build temperature, only measurements from the first three thermocouples (KTC1, KTC2, KTC3) are stacked as a single dataset for training, while the last thermocouple (KTC4) is purely used for testing.

*Parameter settings for AFSD-Physics.* The parameters for human-informed machine learning modeling are set as follows: $M = 1000$, $k \in \{3,4,5\}$, and $\lambda_2 = 100$. The best $k$ value is selected after a preliminary tuning by simulating the acquired model for 5 layers using temperatures from the 21-th layer of the 135-rpm dataset as initial values. For the heat generation stage of tool temperature evolution, the features include the tool temperature $T_{tool}$, spindle speed $\omega$, spindle torque $T_f$, feedstock feed speed $f_m$, and actuator force $F_m$ to push the feedstock downward. Four-order polynomials are used to construct the candidate nonlinear physical terms $\boldsymbol{\theta}(T, \boldsymbol{u})$. After the first round of training, only terms related to $T_{tool}$, $\omega$ and $T_f$ are identified. As such in the second round of training, only these three features are included to learn the governing equation in Eq. (14a). For the tool cooling stage, only $T_{tool}$ is included to learn Eq. (14b) since all process parameters are not related to tool cooling and there are no measurements from the water jacket. To learn Eq. (15) for build temperature evolution, the feature set consists of the tool temperature $T_{tool}$, the build temperature $T_{build}$, and the distance $d$ between a thermocouple in the substrate and the tool.

*Comparison metrics.* In addition to the computational efficiency comparison with simulation time, the mean absolute percentage error (MAPE) in Eq. (16) is used to measure the prediction accuracy of the methods under comparison.

$$MAPE = \frac{1}{N}\sum_{i=1}^{N} \frac{|T_i - \hat{T}_i|}{T_i} \times 100\%, \tag{16}$$

where $T_i$ and $\hat{T}_i$ are the measurement and prediction of the temperature, respectively, and $N$ is the total number of measurements. The smaller the MAPE value the better accuracy of the method.

### 6.2 Results and analysis for tool temperature

For Eq. (14), the coefficients are acquired as $\boldsymbol{a} = [a_1\ a_2\ a_3] = [2.7640 \times 10^{-9}\ \ 1.1382 \times 10^{-8}\ \ 1.8361 \times 10^{-9}]$, $\boldsymbol{b} = [b_1\ b_2\ b_3] = [0.3282\ \ 0.0135\ \ 6.0601 \times 10^{-6}]$. Type I simulations of the acquired governing equations in Eqs. (14a) and (14b) are first presented for heat generation and tool cooling, respectively. Then Type II simulation is conducted to predict the coupled heat generation and tool cooling stages.

*6.2.1 Simulation results for tool heating*

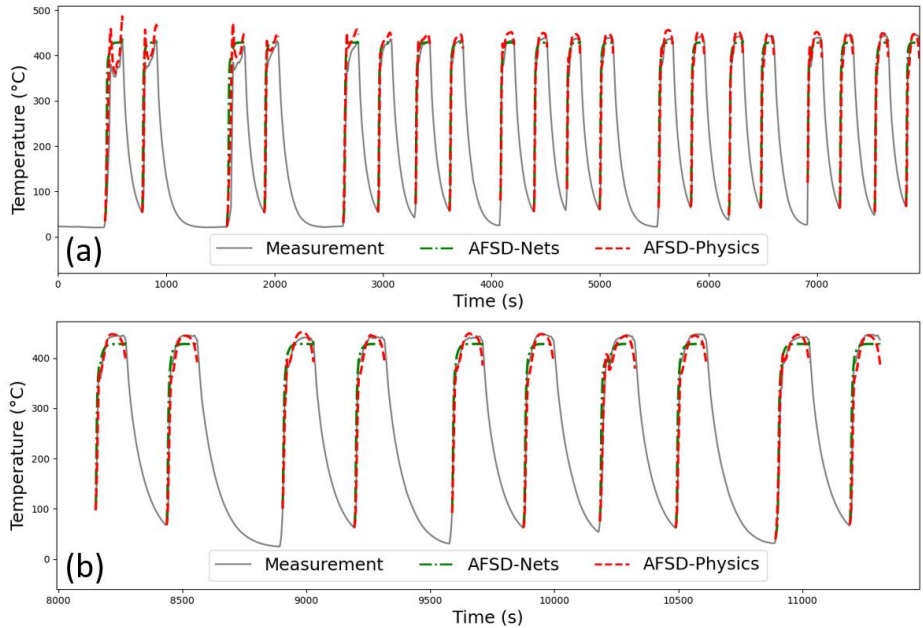

*Figure 3. Type I simulation (a new initial value for each layer) of heat generation on training and testing set of 135-rpm in (a) and (b), respectively.*

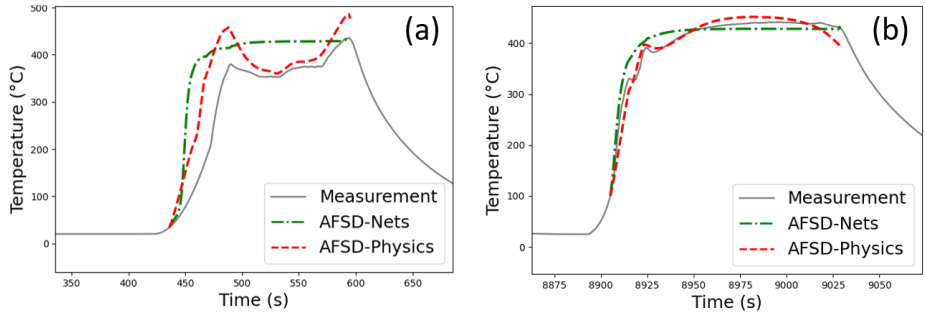

*Figure 4. Zoom-in plots for Figure 3.*

Figure 3 displays the simulation results for the temperature profile of the tool heat generation stage using the training and testing sets in (a) and (b), respectively. The zoom-in plots are shown in Figure 4. From these figures, both AFSD-Physics and AFSD-Nets can capture the overall trend of temperature increase during the tool heat generation stage. It is observed that AFSD-Physics is in good agreement with the measurements and can capture the details of the temperature evolution. For example, AFSD-Physics captures the first peak value when the spindle speed is decreased from 350 rpm (for quick heating and softening of the feedstock) to 135 rpm (for deposition). The feedstock feed velocity is accordingly increased from 0.93 mm/second to 1.93 mm/second, and the tool starts to traverse to deposit materials. The major heat generation source changes from adiabatic heating caused by plastic deformation to frictional heating. On the contrary, the AFSD-Nets model tends to overly smooth the prediction and fails to recognize the valley-shaped heating pattern present in the first 5 layers of the training set, see Figure 4(a). Note that the measurements of the first 5 layers are not included for training, as such the acquired model of AFSD-Physics amplifies valley-shaped heating patterns for the first 5 layers. In general, AFSD-Physics outperforms AFSD-Nets in capturing the overall profile, detailed heating pattern, and peak temperatures for the heat generation stage.

The MAPE and simulation time of the acquired models on the training and testing sets of 135-rpm are presented in Table 1. It is observed that AFSD-Physics obtains much smaller MAPE, significantly outperforming AFSD-Nets for both training and testing datasets. In particular, the MAPE of AFSD-Physics is at most 55% of that of AFSD-Nets. The simulation time of AFSD-Physics is within 0.16 seconds. This indicates that AFSD-Physics can achieve both physically interpretable and robust models with high accuracy and low computational cost.

*Table 1. MAPE and simulation time comparison for Type I simulation (a new initial value for each layer) of heat generation stage.*

| Dataset | AFSD-Nets | | AFSD-Physics | |
|---|---|---|---|---|
| | MAPE (%) | Time (s) | MAPE (%) | Time (s) |
| Training set, 135-rpm | 18.5429 | 0.6882 | **10.2894** | **0.1566** |
| Testing set, 135-rpm | 7.3858 | 0.4169 | **3.7789** | **0.1496** |

*6.2.2 Simulation results for tool cooling*

Figure 5 displays the Type I simulation results for the temperature trajectory comparison of the tool cooling stage using the training and testing sets in (a) and (b), respectively. Both AFSD-Nets and AFSD-Physics accurately capture the temperature decrease during the tool cooling stage and align almost perfectly with the measurements. In fact, the tool cooling stage has a much smoother profile than the heat generation stage, mainly because no process parameters are involved. As the tool leaves contact with the build, the temperature evolution during cooling depends mainly on the conduction along the feedstock and through the cooling jacket, both of which can be considered to occur at a constant rate. As such, the trend of tool cooling is clear and can be more easily captured by both AFSD-Nets and AFSD-Physics.

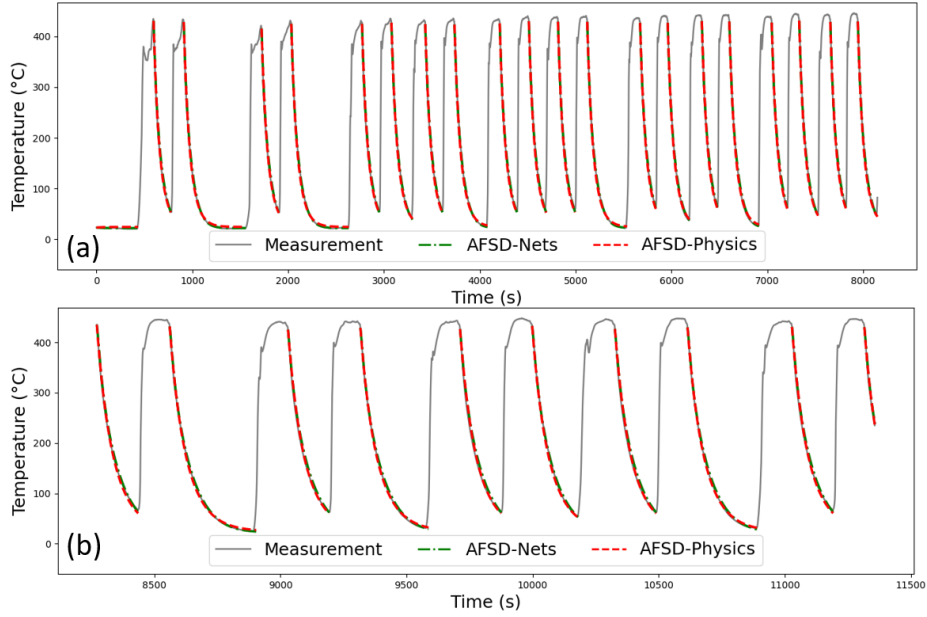

*Figure 5. Type I simulation (a new initial value for each layer) of tool cooling stage on training and testing set of 135-rpm in (a) and (b), respectively.*

Table 2 summarizes the MAPE and simulation time of Type I simulation results for both the AFSD-Nets and AFSD-Physics models. It is observed that AFSD-Nets model obtains better MAPE prediction performance on the training set of the 135-rpm dataset, while the AFSD-Physics model obtains better prediction performance on the testing set of the 135-rpm dataset. This observation indicates that AFSD-Physics has superior generalization ability for unseen data by capturing the underlying physics as analytical physical terms. Furthermore, AFSD-Nets has 17 parameters for the single hidden layer neural network deployed with 4 neurons. In contrast, the AFSD-Physics acquired model demonstrates superior performance on test data despite having only 3 parameters. Table 2: MAPE and simulation time comparison for Type I simulation (a new initial value for each layer) of tool cooling stage.

| Dataset | AFSD-Nets | | AFSD-Physics | |
|---|---|---|---|---|
| | MAPE (%) | Time (s) | MAPE (%) | Time (s) |
| Training set, 135-rpm | **2.9297** | 5.1711 | 5.8811 | **3.3623** |
| Testing set, 135-rpm | 3.825 | 3.5413 | **3.525** | **3.3489** |

*6.2.3 Coupled simulation of tool temperature evolution*

Type II simulation is conducted to couple the heat generation and tool cooling stages for the entire 30-layer AFSD process. As the initial value of the next layer simulation depends on the last value of the previous one, Type II simulation results are more effective to validate the coupling performance of the acquired governing equations.

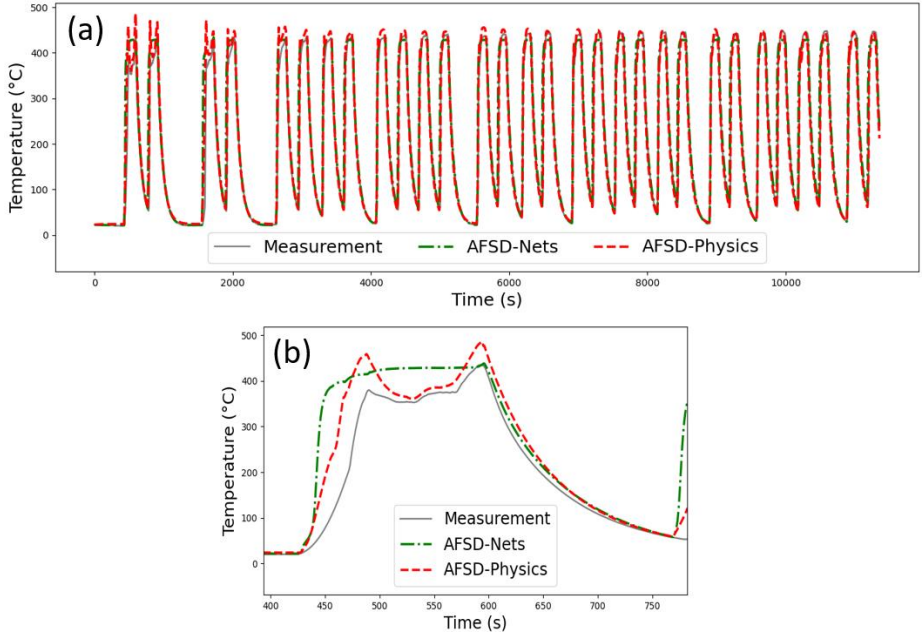

*Figure 6. (a) Type II simulation (a single initial value for the entire multi-layer simulation) of tool temperature on 135-rpm. (b) Zoom-in plot.*

*Table 3. MAPE and simulation time comparison for Type II simulation (a single initial value for the entire multi-layer simulation) of tool temperature.*

| Dataset | AFSD-Nets | | AFSD-Physics | |
|---|---|---|---|---|
| | MAPE (%) | Time (s) | MAPE (%) | Time (s) |
| 135-rpm dataset | 14.5670 | 5.7751 | **9.3273** | **4.2441** |

Figure 6 shows the temperature profile for Type II simulation on the 135-rpm dataset. It is observed that both AFSD-Nets and AFSD-Physics models can simulate the overall thermal cycling. Still, AFSD-Physics can be more effective to capture the valley-shaped heat generation pattern between two peak temperature values. As shown in Table 3, AFSD-Physics outperforms AFSD-Nets for both MAPE and simulation time. Note that about 3 seconds out of the 4.2441 seconds is spent manipulating the data. The simulation for the AFSD-Physics model itself is generally completed within 1.5 seconds for a 30-layer deposition.

**6.3 Results and analysis for build temperature**

For Eq. (15), the coefficients are acquired as $\boldsymbol{c} = [c_1\ c_2\ c_3\ c_4] = [-6.3398 \times 10^{-10}\ 4.6003 \times 10^{-6}\ 2.1208 \times 10^{-7}\ 4.3869 \times 10^{-8}]$. Figure 7 displays the Type II simulation results of the build temperature evolution for the four thermocouples. In general, both AFSD-Nets and AFSD-Physics capture the overall trend of the build temperature profile. Moreover, it is observed that the AFSD-Nets model has better simulation accuracy for the first few layers, but inaccurately predicts the peak temperature with a downward deviation. The AFSD-Physics model has larger deviation in its prediction of the valley value but maintains a strong capability for predicting the peak temperature. It must be emphasized that only the measurements of KTC1, KTC2 and KTC3 are used for training. Therefore, the simulation results of temperature evolution for KTC4 are pure extrapolation. In Figure 7(d), AFSD successfully captures the overall temperature evolution for unseen data, especially for the peak values. This reflects the model's capability to generalize and stabilize.

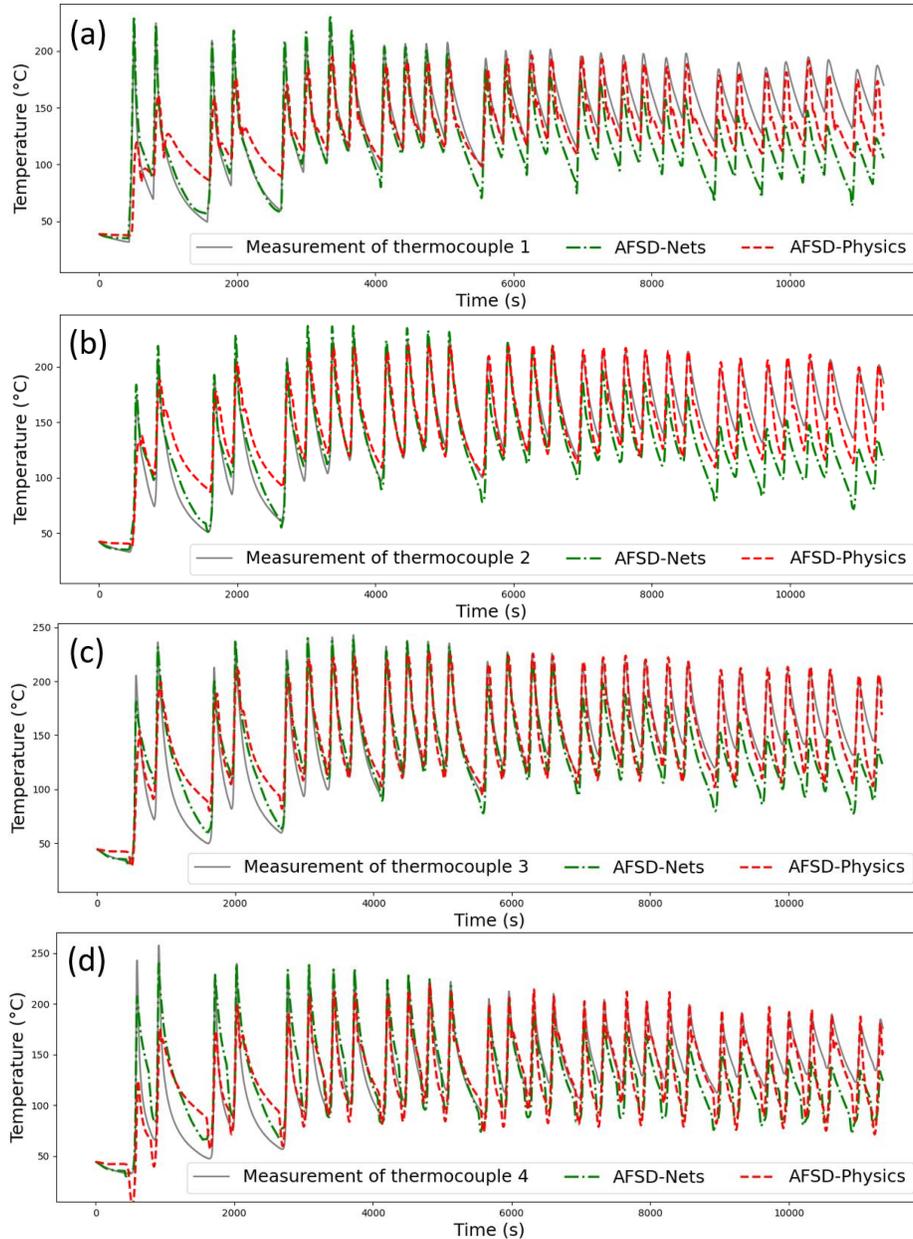

*Figure 7: Type II simulation (a single initial value for the entire multi-layer simulation) of build temperature on 135-rpm.*

Table 4 summarizes the comparison of MAPE and simulation time for build temperature simulation. It is clearly seen that AFSD-Physics outperforms AFSD-Nets for both metrics. The MAPE metrics of build temperature for both methods are worse than those of tool temperature. The main reason for this may lie in the inappropriate use of distance $d$ when the tool is not in contact with the build, in which case the conduction between the tool and the build is terminated and cannot be represented by Euclidean distance on the build geometry. Along with the results in Table 3, the AFSD-Physics acquired governing equations show good agreement with the measurements for the tool and build temperature. The total simulation time is within 5 seconds.

*Table 4. MAPE and simulation time comparison for Type II simulation (a single initial value for the entire multi-layer simulation) of build temperature.*

| Thermocouple | AFSD-Nets | | AFSD-Physics | |
| --- | --- | --- | --- | --- |
| | MAPE (%) | Time (s) | MAPE (%) | Time (s) |
| KTC1 | 17.0176 | 2.3051 | **14.2840** | **0.1177** |
| KTC2 | 14.7449 | 2.3011 | **12.9856** | **0.1057** |
| KTC3 | 15.7802 | 2.4982 | **14.1893** | **0.1087** |
| KTC4 | 19.8777 | 2.1001 | **17.6206** | **0.1044** |

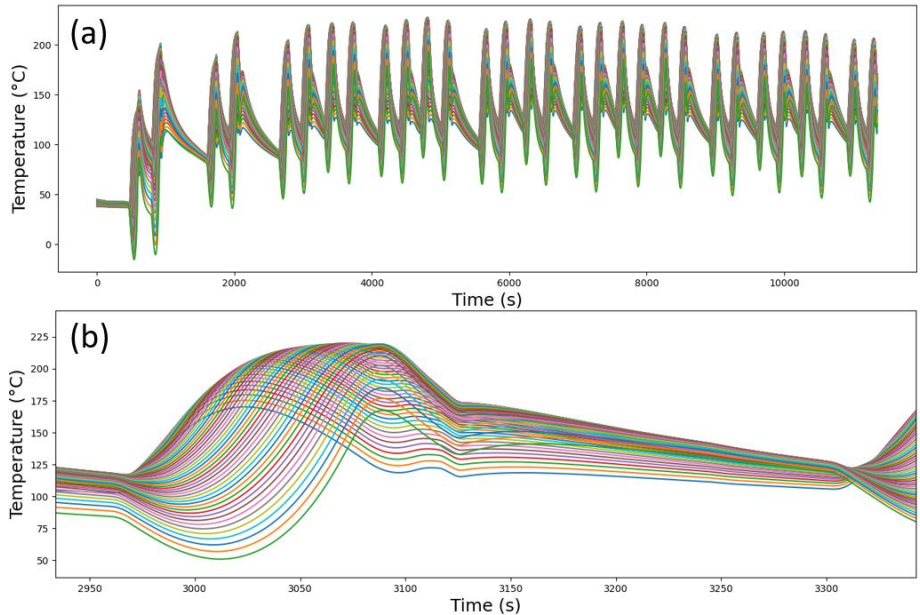

*Figure 8. (a) Simulation results of the build temperature for 53 interpolation locations on the centerline of substrate where four thermocouples are located. (b) Zoom-in plots of one-layer build temperature simulation.*

Finally, the governing equation in Eq. (15) is used to predict the temperature evolution for points on the centerline of the substrate where four thermocouples are located. 53 locations on this centerline are interpolated across the wall length. The Type II simulation results are displayed in Figure 8. This shows the capability of AFSD-Physics model to produce a 1D temperature map. It also shows the potential for 2D/3D temperature maps by integrating more spatial in-process measurements.

## 7 Experimental validation

The proposed AFSD-Physics method is validated using new experiments. The same governing equations from Eq. (14) for tool temperature evolution are acquired using measurements from different deposition settings. For build temperature, governing equations with two out of four physical terms in Eq. (15) were stably acquired. Due to the significance of the tool temperature model and the page limit, results of the tool temperature governing equations are reported here.

In the new experiments, a spindle speed of 115 rpm was used for deposition. Other process parameters including the feedstock feed velocity and tool feed velocity, wall geometry, deposition path, and thermocouples retained the same settings as for deposition at a spindle speed of 135 rpm. A total of 13,782 in-process data points were obtained. The measurements of the first 10 layers include many abnormalities and thus were discarded. The resulting 7782 data points, denoted by 115-rpm, were first used individually and then combined with the 135-rpm dataset to create two sets of new governing equations for tool temperature evolution using AFSD-Physics.

### 7.1 Acquired governing equations

The same set of governing equations are stably acquired for tool temperature using datasets for only 115-rpm and for 135-rpm and 115-rpm combined, as shown in Eq. (17).

$$\dot{T}_{tool} = \begin{cases} a_1\omega^3 T_f + a_2\omega^2 T_f T_{tool} - a_3\omega T_{tool}^3, t \in P_{heat}, & (17a) \\ b_1 - b_2 T_{tool} - b_3 T_{tool}^2, \quad t \in P_{cool}. & (17b) \end{cases}$$

As evidenced by the same physical terms in Eq. (17) as those in Eq. (14), the proposed AFSD-Physics method can stably learn the same set of physically interpretable governing equations. Together with the acquired governing equations using dataset 135-rpm, three settings of parameter values are acquired and presented in Table 5. The changes in parameter values across datasets reveal that these parameters are process dependent. In other words, with different datasets for different deposition settings, the same physical terms are acquired but with different parameter (coefficient) values. This indicates that the acquired governing equations capture the main physical mechanisms for tool-deposition temperature evolution.

Table 5. Parameter values of the acquired governing equations using dataset 135-rpm, 115-rpm, and their combination.

| Parameter | AFSD-Physics: 135 | AFSD-Physics: 115 | AFSD-Physics: 135&115 |
|---|---|---|---|
| $a_1$ | $2.7640 \times 10^{-9}$ | $2.1621 \times 10^{-9}$ | $2.8781 \times 10^{-9}$ |
| $a_2$ | $1.8361 \times 10^{-9}$ | $1.7824 \times 10^{-9}$ | $1.6599 \times 10^{-9}$ |
| $a_3$ | $1.1382 \times 10^{-8}$ | $1.1865 \times 10^{-8}$ | $1.0621 \times 10^{-8}$ |
| $b_1$ | 0.3283 | 0.0198 | 0.1187 |
| $b_2$ | 0.0135 | 0.0044 | 0.0083 |
| $b_3$ | $6.06 \times 10^{-6}$ | $2.7474 \times 10^{-5}$ | $1.8550 \times 10^{-5}$ |

## 7.2 Simulation results and analysis of tool temperature

Both Type I and Type II simulations are conducted for the three models in Eqs. (17) and Table 5. The comparison with AFSD-Nets is not included as the AFSD-Physics acquired model from the 135-rpm dataset significantly outperforms AFSD-Nets, as shown in Section 6. Also, the simulation time of the three models is comparable and thus is not reported in this section.

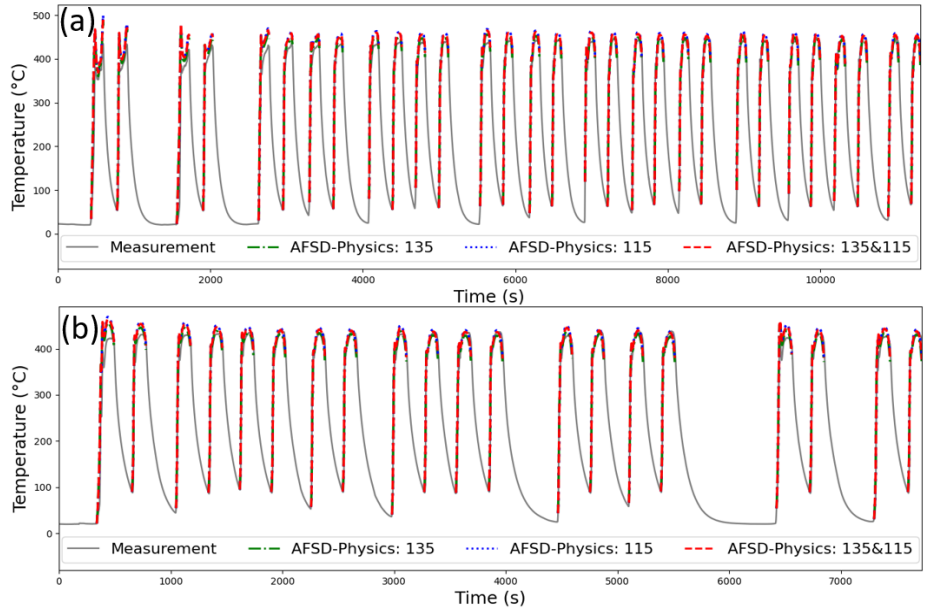

Figure 9. Type I simulation (a new initial value for each layer) of heat generation for 135-rpm in (a) and 115-rpm in (b).

For Type I simulation, Figure 9 and Figure 10 show the result comparisons of heat generation on datasets 135-rpm and 115-rpm and the respective zoom-in plots. Good agreement is observed with the measurements for all three models. Also, all models can capture the detailed valley-shaped heating pattern between any two peak values. In the peak temperature period, the acquired model from the combined 135-rpm and 115-rpm dataset performs somewhere in between the other two models. This may be because each of the other two models overfits its own respective training dataset. Table 6 clearly shows that the model acquired from the 135-rpm dataset obtains the best MAPE performance for 135-rpm, and the same holds true for 115-rpm. The same observations and conclusions can be obtained for the cooling stage, as shown in Figure 11, Figure 12 and Table 7.

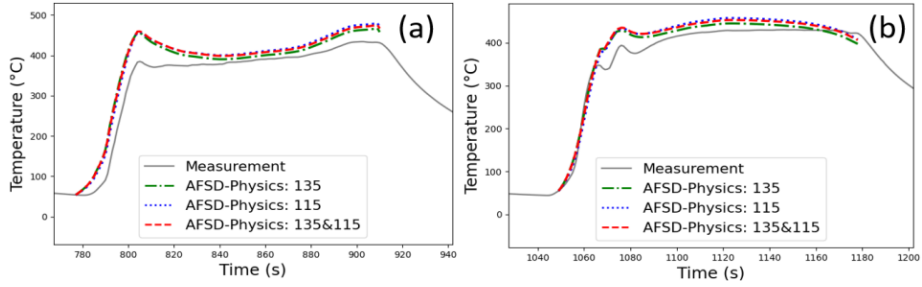

Figure 10. Zoom-in plots for one-layer simulation of Figure 9.

Table 6. MAPE comparison of the AFSD-Physics acquired models for Type I simulation (a new initial value for each layer) of heat generation stage.

| Dataset | AFSD-Physics: 135 | AFSD-Physics: 115 | AFSD-Physics: 135&115 |
| --- | --- | --- | --- |
| 135-rpm | **8.1674** | 8.8862 | 9.0292 |
| 115-rpm | 7.3293 | **7.1952** | 7.4722 |

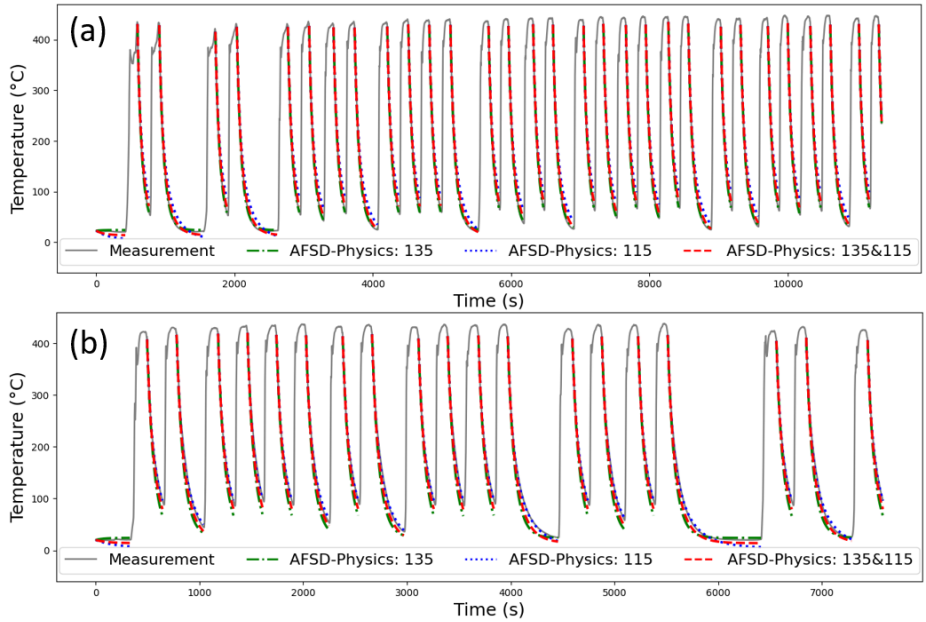

Figure 11. Type I simulation (a new initial value for each layer) for tool cooling stage on dataset 135-rpm in (a) and 115-rpm in (b).

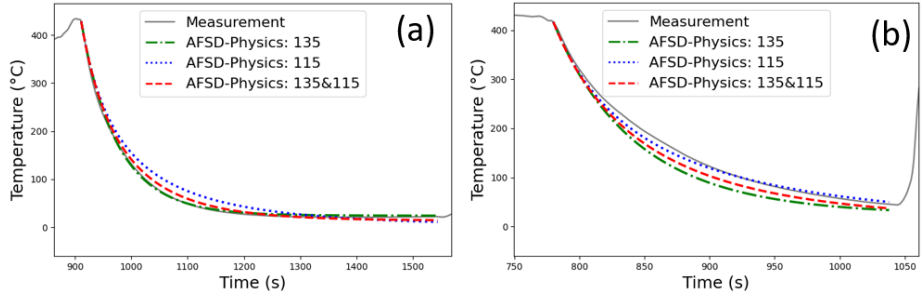

*Figure 12. Zoom-in plots for one-layer simulation of Figure 11.*

*Table 7. MAPE comparison of the AFSD-Physics acquired models for Type I simulation (a new initial value for each layer) of tool cooling stage.*

| Dataset | AFSD-Physics: 135 | AFSD-Physics: 115 | AFSD-Physics: 135&115 |
| --- | --- | --- | --- |
| 135-rpm | **5.2751** | 27.3814 | 11.7532 |
| 115-rpm | 17.4896 | **10.6491** | 14.8682 |

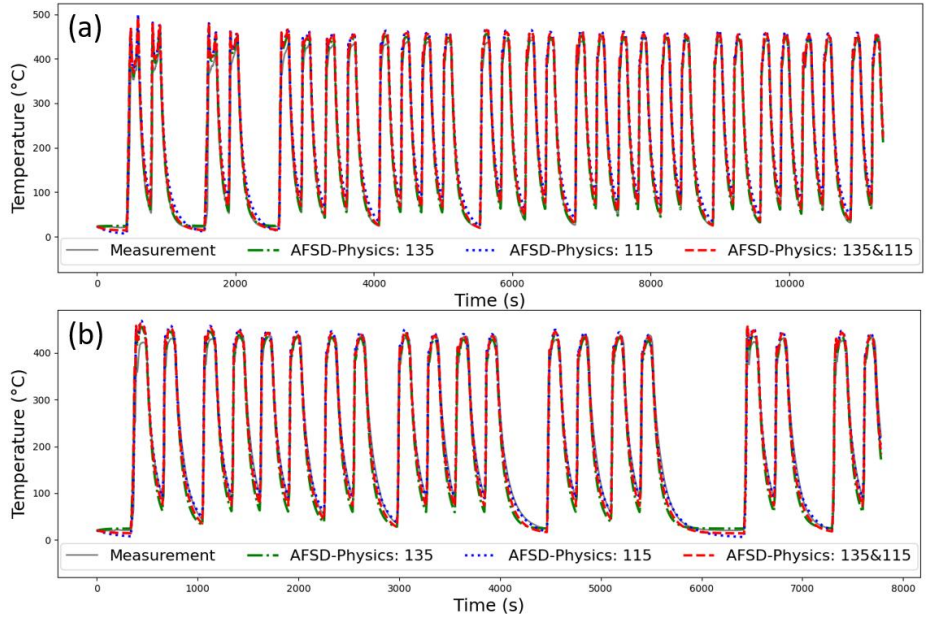

*Figure 13. Type II simulation (a single initial value for the entire multi-layer simulation) of tool temperature on dataset 135-rpm in (a) and 115-rpm in (b).*

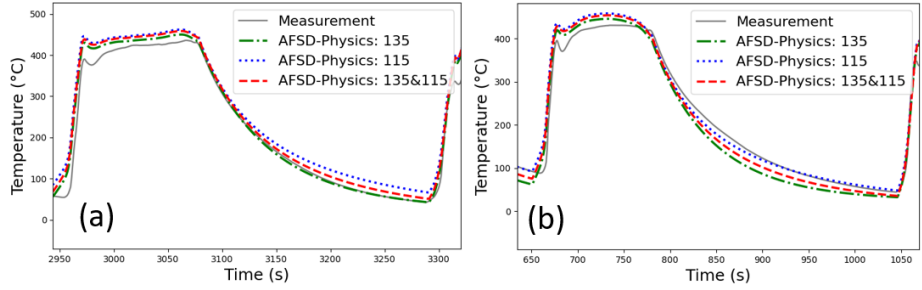

*Figure 14. Zoom-in plots for one-layer simulation of Figure 13.*
*Table 8. MAPE comparison of the AFSD-Physics acquired models for Type II simulation (a single initial value for the entire multi-layer simulation).*

| Dataset | AFSD-Physics: 135 | AFSD-Physics: 115 | AFSD-Physics: 135&115 |
|---|---|---|---|
| 135-rpm | **9.3273** | 23.4318 | 12.7625 |
| 115-rpm | 16.5819 | **11.0977** | 14.2969 |

For Type II simulation, Figure 13 and Figure 14 display the coupled simulation result comparisons for 135-rpm and 115-rpm and the respective zoom-in plots. The corresponding MAPE comparisons are summarized in Table 8. The same conclusions from the individual heat generation and cooling stages appear to hold true for their coupled simulation.

In summary, experimental validation shows that the proposed AFSD-Physics method can stably learn the same governing equations from different AFSD settings. The main physical mechanisms are captured by the identified nonlinear physical terms. This indicates that the AFSD-Physics acquired models are physically interpretable and robust models with high-accuracy and low-cost.

## 8. Conclusions and outlook

This paper presents a modeling effort for the temperature evolution of an emerging solid state additive manufacturing process, additive friction stir deposition (AFSD). A human-AI teaming approach is proposed to explore the governing equations of temperature evolution at the tool and the build during AFSD. The proposed human-AI teaming approach presents a pathway to provide AI with first principles models to advance knowledge in manufacturing. The resulting human-informed machine learning method, denoted as AFSD-Physics, can effectively explore the unknown physics of AFSD and learn the governing equations of temperature evolution from in-process measurements. The acquired governing equations provide physically interpretable robust models with low computational cost and high accuracy. Simulations of the acquired governing equations show good agreement with measurements and significantly outperform the state-of-the-art neural network-based machine learning model. Experimental validation with a new process parameter shows the robustness and generalizability of the acquired governing equations.

This study has two main limitations. The primary one is that only temperatures are treated as state variables with respective governing equations, while other explicit physical variables that can contribute to temperature evolution, like force experienced by the build, are not considered. Another limitation is that only spindle speed is involved as a process parameter in the acquired governing equations. Future work will develop analytical models and integrate them with a series of representative experiments to enable the proposed human-AI teaming approach for exploring governing equations with arbitrary process parameters, including tool spindle speed, feedstock feed velocity, and tool traverse speed. Additionally, governing equations with other physical variables for the build will be explored to advance the convergence of tool-process-structure-property for AFSD.

**Acknowledgements**

The authors acknowledge support from the NSF Engineering Research Center for Hybrid Autonomous Manufacturing Moving from Evolution to Revolution (ERC-HAMMER) under Award Number EEC-2133630. This work was partially supported by the DOE Office of Energy Efficiency and Renewable Energy (EERE), under contract DE-AC05 00OR22725. The authors also acknowledge the seed funding from the AI TENNessee Initiative to partially support this research.

**References**

[1] Mishra, R.S., R.S. Haridas, and P. Agrawal, Friction stir-based additive manufacturing. Science and Technology of Welding and Joining, 2022. 27(3): pp. 141-165.

[2] Yu, H.Z. and R.S. Mishra, Additive friction stir deposition: a deformation processing route to metal additive manufacturing. Materials Research Letters, 2021. 9(2): pp. 71-83.

[3] Jin, Y., T. Yang, T. Wang, S. Dowden, A. Neogi, and N.B. Dahotre, Behavioral simulations and experimental evaluations of stress induced spatial nonuniformity of dynamic bulk modulus in additive friction stir deposited AA 6061. Journal of Manufacturing Processes, 2023. 94: pp. 454-465.

[4] Perry, M.E., H.A. Rauch, R.J. Griffiths, D. Garcia, J.M. Sietins, Y. Zhu, Y. Zhu, and Z.Y. Hang, Tracing plastic deformation path and concurrent grain refinement during additive friction stir deposition. Materialia, 2021. 18: pp. 101159.

[5] Zhu, N., D. Avery, Y. Chen, K. An, J. Jordon, P. Allison, and L. Brewer, Residual stress distributions in aa6061 material produced by additive friction stir deposition. Journal of Materials Engineering and

Performance, 2023. 32(12): pp. 5535-5544.

[6] Perry, M.E., R.J. Griffiths, D. Garcia, J.M. Sietins, Y. Zhu, and Z.Y. Hang, Morphological and microstructural investigation of the non-planar interface formed in solid-state metal additive manufacturing by additive friction stir deposition. Additive Manufacturing, 2020. 35: pp. 101293.

[7] Griffiths, R.J., D. Garcia, J. Song, V.K. Vasudevan, M.A. Steiner, W. Cai, and Z.Y. Hang, Solid-state additive manufacturing of aluminum and copper using additive friction stir deposition: Process-microstructure linkages. Materialia, 2021. 15: pp. 100967.

[8] Williams, M., T. Robinson, C. Williamson, R. Kinser, N. Ashmore, P. Allison, and J. Jordon, Elucidating the effect of additive friction stir deposition on the resulting microstructure and mechanical properties of magnesium alloy we43. Metals, 2021. 11(11): pp. 1739.

[9] Joshi, S.S., S.M. Patil, S. Mazumder, S. Sharma, D.A. Riley, S. Dowden, R. Banerjee, and N.B. Dahotre, Additive friction stir deposition of AZ31B magnesium alloy. Journal of Magnesium and Alloys, 2022. 10(9): pp. 2404-2420.

[10] Beladi, H., E. Farabi, P.D. Hodgson, M.R. Barnett, G.S. Rohrer, and D. Fabijanic, Microstructure evolution of 316L stainless steel during solid-state additive friction stir deposition. Philosophical Magazine, 2022. 102(7): pp. 618-633.

[11] Hartley, W.D., D. Garcia, J.K. Yoder, E. Poczatek, J.H. Forsmark, S.G. Luckey, D.A. Dillard, and Z.Y. Hang, Solid-state cladding on thin automotive sheet metals enabled by additive friction stir deposition. Journal of Materials Processing Technology, 2021. 291: pp. 117045.

[12] Garcia, D., W.D. Hartley, H.A. Rauch, R.J. Griffiths, R. Wang, Z.J. Kong, Y. Zhu, and Z.Y. Hang, In situ investigation into temperature evolution and heat generation during additive friction stir deposition: A comparative study of Cu and Al-Mg-Si. Additive Manufacturing, 2020. 34: pp. 101386.

[13] Zeng, C., H. Ghadimi, H. Ding, S. Nemati, A. Garbie, J. Raush, and S. Guo, Microstructure evolution of Al6061 alloy made by additive friction stir deposition. Materials, 2022. 15(10): pp. 3676.

[14] Phillips, B., C. Mason, S. Beck, D. Avery, K. Doherty, P. Allison, and J. Jordon, Effect of parallel deposition path and interface material flow on resulting microstructure and tensile behavior of Al-Mg-Si alloy fabricated by additive friction stir deposition. Journal of Materials Processing Technology, 2021. 295: pp. 117169.

[15] Kincaid, J., R. Zameroski, T. No, J. Bohling, B. Compton, and T. Schmitz. Hybrid Manufacturing: Combining Additive Friction Stir Deposition, Metrology, and Machining. in TMS Annual Meeting & Exhibition. 2023. pp. 3-13: Springer.

[16] Schmitz, T., L. Costa, B.K. Canfield, J. Kincaid, R. Zameroski, R. Garcia, C. Frederick, A.M. Rossy, and T.M. Moeller, Embedded QR code for part authentication in additive friction stir deposition. Manufacturing Letters, 2023. 35: pp. 16-19.

[17] Kincaid, J., E. Charles, R. Garcia, J. Dvorak, T. No, S. Smith, and T. Schmitz, Process planning for hybrid manufacturing using additive friction stir deposition. Manufacturing Letters, 2023.

[18] Stubblefield, G., K. Fraser, B. Phillips, J. Jordon, and P. Allison, A meshfree computational framework for the numerical simulation of the solid-state additive manufacturing process, additive friction stir-deposition (AFS-D). Materials & Design, 2021. 202: pp. 109514.

[19] Stubblefield, G., K. Fraser, D. Van Iderstine, S. Mujahid, H. Rhee, J. Jordon, and P. Allison, Elucidating the influence of temperature and strain rate on the mechanics of AFS-D through a combined experimental and computational approach. Journal of Materials Processing Technology, 2022. 305: pp. 117593.

[20] Stubblefield, G., K. Fraser, T. Robinson, N. Zhu, R. Kinser, J. Tew, B. Cordle, J. Jordon, and P. Allison, A computational and experimental approach to understanding material flow behavior during additive friction stir deposition (AFSD). Computational Particle Mechanics, 2023: pp. 1-15.

[21] Kincaid, K.C., D.W. MacPhee, G. Stubblefield, J. Jordon, T.W. Rushing, and P. Allison, A finite volume framework for the simulation of additive friction stir deposition. Journal of Engineering Materials and Technology, 2023. 145(3): pp. 031002.

[22] Gotawala, N. and Z.Y. Hang, Material flow path and extreme thermomechanical processing history during additive friction stir deposition. Journal of Manufacturing Processes, 2023. 101: pp. 114-127.

[23] Joshi, S.S., S. Sharma, M. Radhakrishnan, M.V. Pantawane, S.M. Patil, Y. Jin, T. Yang, D.A. Riley, R. Banerjee, and N.B. Dahotre, A multi modal approach to microstructure evolution and mechanical response of additive friction stir deposited AZ31B Mg alloy. Scientific Reports, 2022. 12(1): pp. 13234.

[24] Sharma, S., K.M. Krishna, M. Radhakrishnan, M.V. Pantawane, S.M. Patil, S.S. Joshi, R. Banerjee, and N.B. Dahotre, A pseudo thermo-mechanical model linking process parameters to microstructural evolution in multilayer additive friction stir deposition of magnesium alloy. Materials & Design, 2022. 224: pp. 111412.

[25] Shi, T., J. Wu, M. Ma, E. Charles, and T. Schmitz, AFSD-Nets: A Physics-informed machine learning model for predicting the temperature evolution during additive friction stir deposition. Under review, 2023.

[26] Shi, Z., H. Ma, H. Tran, and G. Zhang, Compressive-sensing-assisted mixed integer optimization for dynamical system discovery with highly noisy data. arXiv preprint arXiv:2209.12663, 2022.

[27] Ma, M., J. Wu, C. Post, T. Shi, J. Yi, T. Schmitz, and H. Wang, A physics-informed machine learning-based control method for nonlinear dynamic systems with highly noisy measurements. ArXiv. /abs/2311.07613, 2023.